**DIA-CLIP: a universal representation learning framework for zero-shot DIA proteomics**

Yucheng Liao[1,2,3], Han Wen[2,3], Weinan E[1,2,4], Weijie Zhang[2,3,*]

[1] Center for Machine Learning Research, Peking University, Beijing, China 100871

[2] AI for Science Institute, Beijing, China 100080

[3] State Key Laboratory of Medical Proteomics, Beijing, China 102206

[4] School of Mathematical Sciences, Peking University, Beijing, China 100871

[*] Corresponding author

**Abstract**

Data-independent acquisition mass spectrometry (DIA-MS) has established itself as a cornerstone of proteomic profiling and large-scale systems biology, offering unparalleled depth and reproducibility. Current DIA analysis frameworks, however, require semi-supervised training within each run for peptide-spectrum match (PSM) re-scoring. This approach is prone to overfitting and lacks generalizability across diverse species and experimental conditions. Here, we present DIA-CLIP, a pre-trained model shifting the DIA analysis paradigm from semi-supervised training to universal cross-modal representation learning. By integrating dual-encoder contrastive learning framework with encoder-decoder architecture, DIA-CLIP establishes a unified cross-modal representation for peptides and corresponding spectral features, achieving high-precision, zero-shot PSM inference. Extensive evaluations across diverse benchmarks demonstrate that DIA-CLIP consistently outperforms state-of-the-art tools, yielding up to a 45% increase in protein identification while achieving a 12% reduction in entrapment identifications. Moreover, DIA-CLIP holds immense potential for diverse practical applications, such as single-cell and spatial proteomics, where its enhanced identification depth facilitates the discovery of novel biomarkers and the elucidates of intricate cellular mechanisms.

**Introduction**

Driven by the burgeoning demands of precision diagnostics and therapeutic development, proteomics has transitioned from a specialized tool to a high-throughput engine for discovery [1][2]. The ability to comprehensively characterize the proteome across vast population cohorts is now pivotal for identifying diagnostic biomarkers, monitoring treatment responses, and elucidating complex biological mechanisms[3][4][5][6]. To meet these rigorous requirements for depth and scalability, Data-independent acquisition mass spectrometry has emerged as a cornerstone of high-throughput proteomics[7][8][9][10], lauded for its high reproducibility[11], sensitivity[12][13], and high-throughput[14]. However, fully realizing the potential of DIA-MS is often hampered by the inherent complexity of its signal deconvolution, which remains a preeminent challenge in computational proteomics[15][16]. Unlike traditional acquisition modes, DIA-MS employs isolation windows designed to encompass multiple m/z ranges, resulting in the concurrent fragmentation of multiple co-eluting precursor ions. This multiplexing gives rise to highly convoluted spectra where fragment-ion signals from disparate peptides are extensively interleaved, thereby confounding the definitive assignment of peptide-spectrum matches (PSMs) and limiting the depth of proteomic coverage[7][17].

To analysis DIA-MS data, peptide-centric analysis pipelines, such as OpenSWATH[18], Spectronaut[19], DIA-NN[20], and MaxDIA[21], typically rely on pre-defined spectral libraries to query DIA-MS data. These workflows conventionally comprise a standardized computational workflow: retention time (RT) calibration, extracted ion chromatogram (XIC) generation, peak-group scoring, PSM re-scoring and false discovery rate (FDR) estimation. To distinguish true target peptides from decoy hits, earlier PSM re-scoring strategies relied heavily on semi-supervised learning algorithms (e.g., Percolator[22], mProphet[23] or XGBoost[24]). These algorithms are underpinned by sophisticated feature engineering, which integrates diverse statistical metrics such as RT alignment, peak-shape correlation, and ionic intensity distributions to enhance identification confidence.

The rise of deep learning has catalyzed the development of advanced re-scoring architectures, such as DreamDIA[25], Alpha-XIC[26], and Alpha-Tri[27], which integrate deep neural networks like Long Short-Term Memory (LSTM) network[28] to evaluate PSMs. Furthermore, DIA-BERT[29] harnesses pre-trained transformer-based architectures to bolster re-scoring performance. Despite these advancements, existing frameworks necessitate the optimization of independent scoring models for individual DIA-MS analysis runs. This 'run-specific' training paradigm harbors an inherent risk of overfitting due to constrained local sample sizes, which often precludes the model from capturing robust, global features across diverse experiments. Moreover, contemporary scoring metrics frequently analyze peak groups in isolation [26][27][28], failing to capture the complex, non-linear semantic associations between the amino acid sequence and multidimensional spectral data. These limitations highlight a pivotal opportunity for a framework capable of precise PSM identification within highly sparse, high-dimensional MS space in a zero-shot manner, obviating the requirement for repetitive optimization .

To circumvent these long-standing bottlenecks, we present DIA-CLIP (Data-Independent Acquisition with Contrastive Learning Integrated Proteomics), a unified, end-to-end model for PSM re-scoring. Diverging from traditional methodologies, DIA-CLIP represents the first implementation of cross-modal contrastive learning in DIA-MS analysis. At its core, DIA-CLIP employs a hybrid architectural design that synergizes a dual-encoder contrastive learning framework with a sophisticated encoder-decoder architecture. The dual-encoder component, which integrates transformer-based sequence encoder and a specialized spectral encoder, serves to align peptide sequences and XIC signals in a shared latent space. Concurrently, a decoder architecture leverages the aligned latent features to extract high-dimensional semantic abstractions. By harnessing extensive pre-training on heterogeneous, large-scale datasets, DIA-CLIP achieves precise PSM identification via zero-shot inference. This approach obviates the necessity for iterative semi-supervised calibrations, thereby establishing a highly generalizable paradigm for proteomic discovery across diverse experimental data.

Furthermore, to demonstrate the broad generalizability of DIA-CLIP, we extensively evaluated DIA-CLIP across broad and heterogeneous DIA-MS proteomics cohorts, encompassing HeLa cell lysates, complex multi-species mixture, clinical breast cancer specimens, and ultra-low-input single-cell preparations. Our evaluations demonstrate a marked expansion in detectable proteome depth and refinement of quantitative accuracy. Specifically, DIA-CLIP achieved up to a 45% increase in protein identifications compared to existing software tools, while concurrently reducing about 12% entrapment-based false identifications. Beyond achieving superior benchmarks, DIA-CLIP demonstrates exceptional translational utility in challenging, practical scenarios, where the zero-shot capability ensures high-performance inference without the need for specific re-training. By enabling a more comprehensive and precise interrogation of the proteome, DIA-CLIP provides a transformative computational foundation for uncovering of previously inaccessible biological insights, facilitating the discovery of novel biomarkers and the elucidation of intricate cellular mechanisms.

**Results**

**Architecture and Identification Workflow of DIA-CLIP**

DIA-CLIP features a unified end-to-end architecture synergizing cross-modal contrastive learning framework [30] with a dual-architectural paradigm comprising both dual-encoder and encoder-decoder architecture (Supplementary Fig. S1). The design rationale centers on the functional decoupling of cross-modal alignment and high-dimensional feature refinement. Specifically, the dual-encoder component utilizes a transformer-based sequence encoder and a specialized spectral encoder to project peptide sequences and XIC signals into a shared latent manifold for establishing foundational semantic correspondences. Complementing this global alignment, the encoder-decoder architecture serves as a discriminative engine, engineered to decode the intricate, non-linear dependencies between peptide structures and spectral signatures. This synergistic design couples the broad generalization of the dual-encoder with the high-resolution sensitivity of the encoder-decoder, facilitating precise PSM adjudication across heterogeneous datasets.

To circumvent the 'run-specific' training constraints inherent in conventional DIA analysis, DIA-CLIP employs a supervised contrastive learning pre-training strategy to internalize robust representations of peptide-spectrum correspondences (Fig. 1a). We first curated a training dataset by processing diverse, multi-species DIA-MS data through DIA-NN, yielding an extensive collection of over 28 million high-confidence PSMs across varying instrumentation platforms (e.g. Astral, TripleTOF). To bolster the discriminative performance of DIA-CLIP against deceptive signals, entrapment PSMs is incorporated as negative sample samples during pre-training. This approach compels DIA-CLIP to learn the subtle spectral nuances required to distinguish true targets from entrapments, ensuring accurate identification in a zero-shot manner.

By pre-training on the heterogeneous, large-scale datasets, DIA-CLIP achieves a robust alignment of precursor characteristics and XIC signals within a shared latent space. This approach fundamentally transcend the constraints of traditional feature engineering, capturing the intrinsic peptide-spectrum correspondences. Crucially, this pre-training paradigm empowers DIA-CLIP with zero-shot inference capabilities, enabling deeper and more accurate PSM identification without the need for laborious finetuning or semi-supervised training on new datasets.

Leveraging the DIA-CLIP, we established a comprehensive, peptide-centric workflow for end-to-end DIA-MS data analysis (Fig. 1b). This pipeline is designed to be software-agnostic. Retention time (RT) calibration can be performed by any conventional DIA identification platform, such as DIA-NN. Subsequently, DIA-CLIP extracts precursor and fragment XICs, performing PSM re-scoring and quantification within a strictly zero-shot scenario. The core strength of this workflow stems from the seamless integration of extensive global prior knowledge, internalized during large-scale pre-training, directly into the analysis pipeline. By imbuing the standardized workflow with these prior knowledge, DIA-CLIP transcends biases inherent in individual experimental runs to ensure consistent, high-fidelity identification across varying experimental conditions.

Detailed descriptions of the dataset generation, training procedures, the re-scoring and quantification algorithms are provided in the Methods.

**Assessment of Identification Performance across Diverse Analytical Landscapes**

To systematically evaluate the efficacy of DIA-CLIP in proteome identification, we initially conducted rigorous benchmarking using HeLa cell lysates DIA-MS data across varying liquid-chromatographic (LC) gradients (PRIDE ID: PXD005573; Fig. 2a) [31]. Furthermore, as the advent of Orbitrap Astral MS instrumentation has fundamentally redefined the landscape of DIA proteomics by substantially increasing spectral density and complexity, we extended our evaluation to a highly multiplexed, multi-species consortium DIA-MS data acquired on this next-generation platform (PRIDE ID: PXD046444; Fig. 3a). This multi-species cohort consisted of proteomes from *Homo sapiens* (human), *Escherichia coli (E. coli),* and *Saccharomyces cerevisiae* (yeast) combined at six distinct mixing ratios. This two-tiered approach allowed us to assess the robustness of DIA-CLIP ranging from standard benchmarks to the heightened complexity of next-generation data. Crucially, to guarantee an unbiased assessment of zero-shot inference capability, all mass spectrometry files used for validation were strictly excluded from the pre-training dataset.

Evaluations of Hela cell lysate across diverse LC gradient(30–240 min; Fig. 2c,d) revealed that the number of peptides and proteins identified by DIA-CLIP markedly surpassed the benchmarks established by existing tools, such as DIA-NN, MaxDIA, and MSFragger-DIA [32]. Taking the 90-minute LC gradient as a representative example, DIA-CLIP achieved a 6.5% and 3.7% increase in peptide and protein identifications, respectively, while maintaining stringent FDR control (Fig. 2b). Visualization of target and decoy PSM score distributions (Supplementary Fig. S2a), further indicated that DIA-CLIP constructs highly discriminative representations. Consensus analysis via Venn diagrams (Fig. 2e) further illustrated that DIA-CLIP not only encompasses the majority of proteins identified by other tools but also recovers unique peptides and proteins. Besides, analysis of peptide sequence length distributions (Supplementary Fig. S2b) confirmed that identifications of DIA-CLIP across various length intervals follow a distribution trend consistent with conventional software tools. The universal increase in identification counts across all length intervals suggested that the performance of DIA-CLIP was free from sequence-length bias.

The massive spectral throughput of the multi-species Astral dataset, characterized by narrow-window sampling and compressed chromatographic timescales, provided a rigorous challenge for identification stability. Within this demanding acquisition environment, DIA-CLIP identified the highest number of valid precursors (Fig. 3b), achieving 1% increase in precursor identifications within 1% FDR compared to DIA-NN. Furthermore, the assessment of identification counts across varying coefficients of variation (CV) for both precursors (Fig. 3c) and proteins (Fig. 3d) demonstrated that DIA-CLIP maintained a substantial lead over existing tools. Notably, in the high-precision regime (CV < 5%), DIA-CLIP yielded approximately triple the number of precursor identifications and double the protein identifications relative to DIA-NN.

Collectively, above results demonstrate that DIA-CLIP surmounts the sensitivity bottlenecks inherent in heuristic feature engineering by leveraging the internalized, cross-modal representations to effectively disentangle authentic signals from intricate spectral scans. This approach establishes a robust computational foundation for high-throughput proteomics, facilitating deeper interrogation of the proteome across the latest generation of mass spectrometry platforms.

**Multi-dimensional Assessment of PSM Assignment Reliability**

Beyond the substantial gains in identification depth, the expanded proteome coverage provided by DIA-CLIP was validated to ensure the authenticity of peptide signals and the exclusion of false-positive noise. To assess the authenticity of precursor ions identified exclusively by DIA-CLIP, we inspected XICs of representative precursor ions that remained undetected by established tools. For instance, in the 90-min HeLa cell lysates analysis, the +2 charged peptide EINYILR was uniquely identified by DIA-CLIP, displaying exceptional peak symmetry, stringent fragment-ion co-elution profiles, and robust signal-to-noise ratios. This qualitative signal integrity is further corroborated by extensive additional XIC examples uniquely recovered by DIA-CLIP across varying gradients (Supplementary Fig. S3).

Given the formidable challenge of accurate error control within DIA workflows [33] and the failure of conventional search tools to maintain consistent FDR thresholds through entrapment procedures [34][35], an entrapment experiment was conducted to rigorously validate the reliability and robustness of DIA-CLIP. This methodology utilized a hybrid multi-species entrapment library comprising protein sequences from Homo sapiens, Saccharomyces cerevisiae, Escherichia coli, and Mus musculus to the analysis of HeLa cell lysates DIA-MS (PRIDE ID: PXD005573; Fig. 2a), with any identified non-human peptides classified as entrapments. As illustrated in Fig. 2g, while maintaining a decoy FDR below 0.01, DIA-CLIP yielded entrapment proportions superior to existing analytical tools across varying LC gradient lengths. Notably, application of more stringent decoy FDR thresholds, specifically 0.005 and 0.001, revealed a marked reduction in entrapment FDR for DIA-CLIP relative to other software (Supplementary Fig. S2e, f). Specifically, under the 60-minute LC gradient, DIA-CLIP achieved a 29.7% reduction in entrapment counts at a 0.001 decoy FDR. In-depth investigation into the sensitivity toward decoy versus entrapment sequences further clarified this superior performance. Analysis of the 60-min gradient data revealed the highest sensitivity toward entrapment precursors in DIA-CLIP, characterized by the fewest falsely identified entrapment ions at any given decoy count (Fig. 2h) and an average 11.88% reduction in entrapments.

Integrated quantitative evaluation across multiple datasets further substantiated the high accuracy of identified signals yielded by DIA-CLIP. In the multi-species consortium with predefined mixing ratios (Yeast: Human: *E. coli* is 5:50:45 and 45:50:5), DIA-CLIP maintained superior quantitative fidelity at both the precursor and protein levels (Fig. 3e–h). Scatter plots of log-transformed relative intensities at both the precursor (Fig. 3e-f) and protein (Fig. 3g-h) levels showed high concordance between experimental ratios quantified by DIA-CLIP and the theoretical expectations (indicated by dashed lines: green for human, blue for yeast, and orange for *E. coli*). Subsequent box plot analysis (Fig. 3i, j) corroborated this performance, with DIA-CLIP exhibiting significantly narrower distributions while maintaining median values comparable to DIA-NN in alignment with the ground truth. This systematic precision is further mirrored in the analysis of unique identifications in HeLa dataset, where the quantitative results for precursors (Supplementary Fig. S2c) and proteins (Supplementary Fig. S2d) identified solely by DIA-CLIP exhibited significantly tighter clustering relative to DIA-NN.

These multi-dimensional evaluations collectively demonstrate the capacity of **DIA-CLIP** for substantial increases in identification depth while maintaining high quantitative and statistical fidelity in zero-shot inference regimes. By effectively resolving the divergence between decoy and entrapment-based error estimates, DIA-CLIP introduces a robust methodological advancement for accurate proteome interrogation. The resulting synergy between expanded coverage and superior precision establishes DIA-CLIP as a foundational tool for high-fidelity, high-throughput proteomics across the latest generation of mass spectrometry platforms.

**Decipher of Spatially Resolved Proteomes via DIA-CLIP**

Given the capacity to delineate protein profiles with precise spatial context, spatial proteomics has emerged as a transformative tool in various fields like tumor microenvironment research and clinical precision medicine[37][38][39]. To evaluate the practicality of DIA-CLIP in complex biological scenarios, a spatially resolved proteomics dataset generated by integrating laser capture microdissection (LCM) with LC-MS is utilized to profile the proteomic landscape of breast cancer tissue sections (iProX ID: PXD045687; Fig. 4a) [40]. Pathological annotations derived from hematoxylin and eosin (H&E) stained images (Fig. 4b) allowed for the precise resolution of spatial information across distinct pathological regions.

Comparative identification assessments across distinct pathological regions (Fig. 4c), demonstrated a significant increasement in protein identification counts across all regions via DIA-CLIP, providing a comprehensive molecular foundation for high-resolution spatial proteomic maps and biomarker discovery. To bolster data reliability, the analysis considered only proteins identified in at least 50% of tissue sections within a given pathological region. Subsequent analysis of key breast cancer classification markers, including HER2/ERBB2 and ER/ESR1 [41], revealed pronounced spatial heterogeneity (Fig. 4d-e). The observed quantitative trends and spatial distributions aligned closely with the H&E-based pathological annotations, enabling the accurate stratification of two tumor subtypes: Tumor 1 (HER2+, ER-, PR-) and Tumor 2 (HER2+, ER+, PR+). These classifications showed high consistency with the findings of the original study [42][43].

The enhanced identification depth provided by DIA-CLIP facilitated the discovery of novel marker proteins through comparative analysis. Volcano plot (Fig. 4f) and Venn diagram (Supplementary Fig. S4a) analyses of differentially expressed proteins (DEPs) between two tumor regions yielded a 6% increase in the identification of significant markers by DIA-CLIP relative to DIA-NN. For instance, Amine oxidase [flavin-containing] A (AOFA), a protein uniquely identified by DIA-CLIP, displayed specific expression within Tumor 1 regions (Fig. 4g). Previous research associates AOFA with highly invasive phenotypes and the epithelial-mesenchymal transition process in breast cancer [44][45], aligning with the aggressive pathological characteristics in Fig. 4b. Numerous additional examples of DEPs uniquely recovered and spatially visualized by DIA-CLIP are provided in the Supplementary Information

(Supplementary Fig. S4b).

Collectively, these results demonstrate the robust applicability of DIA-CLIP for the accurate interrogation of spatially resolved clinical proteomes. Beyond providing substantial gains in identification counts, the successful stratification of tumor subtypes and the discovery of novel, functionally pertinent protein markers underscore the capacity of DIA-CLIP for deep biological discovery.

**Sensitive Protein Profiling at Single-Cell Resolution**

Single-cell mass spectrometry-based proteomics is hindered by formidable technical bottlenecks, including ultra-low sample input, extreme signal sparsity, and pervasive data missingness [13][47][48][49][50]. Evaluation of DIA-CLIP in these ultra-low input scenarios involved benchmarking against a dataset consisting of 12 technical replicates of HeLa single-cell samples, acquired with 2-Th isolation windows and 3-ms maxITs (PRIDE ID: PXD049211) [51].

Across all replicates, DIA-CLIP consistently and significantly outperformed other software tools in the identification of both precursors (Fig. 5a) and proteins (Fig. 5b). In addition, an in-depth assessment of quantitative stability revealed a substantial increase in the number of precursors (Fig. 5c) and proteins (Fig. 5d) identified by DIA-CLIP with various CV. Furthermore, validation of the expanded single-cell proteome involved inspecting XICs of representative precursors unique to the DIA-CLIP analysis. At single-cell resolution, chromatographic signals frequently deviate from ideal profiles, characterized by severe peak tailing or intermittent signal gaps resulting from ultra-low ion counts (Fig. 5f, g). Despite these artifacts, DIA-CLIP accurately resolved the corresponding signals, effectively surmounting the sensitivity limits of conventional heuristic-based algorithms in high-noise environments. This superior recovery of limit-of-detection ion signals translated directly into enhanced data completeness, a critical metric for single-cell analytical performance. Comparative analysis of completeness heatmaps (Fig. 5e) revealed a significant reduction in missing values with DIA-CLIP, facilitating a more coherent protein detection profile and improved sensitivity toward low-abundance proteins across replicates. The resulting systemic consistency was further corroborated by t-SNE clustering analysis [52], which yielded more compact grouping of replicates for DIA-CLIP (Fig. 5h). Specifically, the mean Euclidean distance in t-SNE space for DIA-CLIP reached 41.015, representing a marked improvement over the 51.366 observed for DIA-NN.

Collectively, the superior identification depth and data completeness achieved by DIA-CLIP establish a robust computational foundation for high-fidelity single-cell proteomics, facilitating the elucidation of subtle biological heterogeneity within signal-sparse experimental landscapes.

**Discussion**

This study introduces DIA-CLIP, a cross-modal model redefining the identification paradigm in data-independent acquisition proteomics by transitioning from semi-supervised training models to a unified, global zero-shot inference architecture. This approach obviates the requirement for repetitive optimization and eliminates the inherent risk of overfitting associated with constrained local sample sizes, representing a significant shift toward generalizable proteomics informatics. The core innovation of DIA-CLIP involves the first implementation of cross-modal contrastive learning within the DIA-MS domain. By synergizing a dual-encoder framework with a sophisticated encoder-decoder architecture, DIA-CLIP aligns peptide sequences and multi-dimensional XIC signals within a shared latent space. This alignment facilitates the capture of complex, non-linear semantic associations between amino acid sequences and spectral data, a capability frequently lacking in contemporary scoring metrics analyzing peak groups in isolation. Furthermore, the utilization of global prior knowledge acquired from millions of PSMs ensures precise zero-shot identification and exceptional robustness across diverse instrumental platforms and biological scenarios.

Systematic benchmarking across heterogeneous cohorts, including complex multi-species consortium, clinical breast cancer specimens and ultra-low-input single-cell preparations, demonstrates a marked expansion in both the depth and accuracy of the detectable proteome. Architecturally, DIA-CLIP permits integration with various peptide-centric search engines, including MaxDIA, Spectronaut, and DIA-NN, for the purpose of PSM re-scoring, further enhancing the utility for generalized re-scoring across diverse computational workflows. Besides, the inference-only nature of DIA-CLIP allows for flexible deployment across both CPU and GPU hardware, facilitating rapid integration into existing computational pipelines without the heavy computational overhead of training or finetuning.

Although the conventional 1% FDR threshold serves as a standard benchmark, it fails to represent an absolute demarcation between authentic and spurious identifications. Due to inherent statistical fluctuations, there remain some authentic target signals in proximity to this threshold. Integrating expert experience to validate identifications within this critical boundary provides a 'human in loop' framework for further elevating the performance ceiling of DIA-CLIP. Establishing a closed-loop system where manually verified high-confidence identifications are fed back into a reinforcement learning framework is capable of facilitating the continuous refinement of the re-scoring model. This adaptive paradigm could transcend conservative FDR constraints to recover authentic signals, effectively pushing the sensitivity limits of DIA-CLIP and enabling the discovery of novel biological features in deep-proteome profiles. Furthermore, expanding the training repertoire to encompass diverse post-translational modifications, non-tryptic peptides, ultraviolet photodissociation (UVPD) fragmentations [53][54] and ion mobility spectrometry (IMS) dimensions will be essential to enhance the versatility of DIA-CLIP across complex, non-standard experimental landscapes.

## Methods

### Mass Spectrometry Files Used in Pre-training

Raw mass spectrometry data utilized for training DIA-CLIP were retrieved from the PRIDE database. Detailed information of these mass spectrometry files is provided in Supplementary Table 1.

### Preparation of the Training Dataset

Files were analyzed for precursor identification using DIA-NN (version 1.7.12; downloaded from https://github.com/YuAirLab/Alpha-Tri). The library-based identification procedure was performed by DIA-NN (version 1.7.12) with parameters '--qvalue 0.01 --matrices --unimod4 --rt-profiling'. Specifically, top-10 fragment ions were selected during the identification procedure. Prediction of the spectral library utilized a separate execution of DIA-NN (version 1.8.1) with parameters '--qvalue 0.01 --matrices --gen-spec-lib --predictor --fasta-search --min-fr-mz 200 --max-fr-mz 1800 --met-excision --cut K*,R* --missed-cleavages 1 --min-pep-len 7 --max-pep-len 30 --min-pr-mz 300 --max-pr-mz 1800 --min-pr-charge 1 --max-pr-charge 4 --unimod4 --reanalyse --relaxed-prot-inf --smart-profiling --peak-center --no-ifs-removal'.

Target precursors identified by DIA-NN (version 1.7.12) served as positive samples, with all decoy precursors functioning as initial negative samples. To expand the negative sample space and enhance model robustness, an entrapment identification procedure was employed. All entrapment precursor identified by DIA-NN (version 1.7.12) were incorporated as additional negative samples.

Extraction of XIC data utilized calibrated retention times and theoretical m/z value of both precursor ions and fragment ions, derived from DIA-NN (version 1.7.12). Precursor ion XICs originated from MS1 spectra, encompassing the monoisotopic precursor m/z and its three associated isotopic peaks. Fragment ion XICs were extracted from MS/MS spectra based on fragment m/z values. Peak groups searching employed a fixed m/z tolerance window of [ion m/z − 30 ppm, ion m/z + 30 ppm]. Within this window, only the peak with the m/z value closest to the theoretical expectation was selected. In instances where no peak was detected within the defined tolerance, the intensity was set to zero and the m/z error was recorded as 30 ppm. Following XIC extraction, both precursor and fragment XICs underwent intensity normalization and interpolation to a fixed dimension of 12 data points to standardize peak shape representation. Finally, the resulting peak groups were processed using a Gaussian smoothing algorithm to minimize signal noise.

### Re-scoring and Quantification Procedure of DIA-CLIP

As illustrated in Figure S1, DIA-CLIP is composed of three primary functional components: initial encoding phase, contrastive learning alignment stage, and specialized decoder for PSM re-scoring.

The initial encoding phase utilizes a dual-encoder architecture. Within a training batch of size B, precursor ions undergo transformation into feature representations $P_1, P_2, ..., P_B$. Besides, to handle the intrinsic relationship between precursor XICs and fragment XICs, DIA-CLIP employs dedicated XIC encoding layers to separately process precursor XICs, fragment XICs, and concatenated XICs. These resulting features are then combined and processed to a representation of spectrum features $S_1, S_2, ..., S_B$.

The secondary stage involves a contrastive learning framework designed to align precursor and spectrum features within a unified, shared latent space. Optimization of DIA-CLIP focuses on maximizing the similarity between $(P_i, S_i)$ with label $y_i = 1$ (positive pairs) while simultaneously minimizing the similarity of $(P_j, S_j)$ with label $y_j = 0$ (negtive pairs).

Derivation of the PSM score utilizes a specialized decoder architecture. Encoded precursor and spectrum features are fed into a transformer-based decoder to generate a [B, dim_model] feature. In parallel, physical co-elution characteristics are explicitly incorporated through the calculation of Pearson Correlation Coefficient (PCC) matrix, subsequently projected into a [B, dim_model] feature via a multi-layer perceptron (MLP). Fusion of these two features occurs through element-wise summation, followed by a final MLP with a sigmoid activation function. This process yields a calibrated PSM score ranging from 0 to 1.

Calculation of quantitative abundance for each identified precursor relies on the arithmetic mean of Top-K areas of fragment XICs. While K conventionally equals 6, DIA-CLIP adaptively selects the top-6 or all available fragment ions with non-zero areas, ensuring the inclusion of only informative signals in the final quantitative result.

**Pre-training Parameters of DIA-CLIP**

DIA-CLIP was jointly pre-trained using the AdamW optimizer with a weight decay coefficient of $1 \times 10^{-6}$. To ensure stable convergence, a cosine annealing scheduler was implemented to dynamically modulate the learning rate, decaying from an initial value of $1 \times 10^{-4}$ to a minimum of $1 \times 10^{-7}$. The training spanned 40 epochs with a batch size of 4096. Dataset was partitioned into training and validation sets at a 9:1 ratio.

**FDR estimation**

To evaluate the statistical confidence of PSMs, we employed the target-decoy strategy. In DIA-CLIP, FDR control followed the Benjamini-Hochberg procedure [55][56]. Specifically, PSMs underwent initial ranking in ascending order of their scores to calculate raw FDR estimates. To ensure the monotonicity of FDR estimates, an adjustment was applied from the beginning of the ranked list, defined as:

$$FDR_{adj}(i) = \max_{i \geq j}\left(\frac{N_{decoy,j}}{N_{target,j} + N_{decoy,j}}\right).$$

Target PSMs with an adjusted FDR below a predefined threshold (e.g., 0.01) were considered statistically significant identifications.

**Benchmark setting**

The benchmarking process included a comparative evaluation against several software tools, specifically DIA-NN (v1.7.12), MaxDIA (v2.6.7.0), MSFragger-DIA (v4.1). Reference proteomes for *Homo sapiens*, *Saccharomyces cerevisiae*, *Escherichia coli*, and *Mus musculus* were retrieved from the UniProt database, while the three-species mixture FASTA file was obtained from the pride repository (PXD002952).

For library-based search workflows, DIA-NN utilized spectral libraries predicted by DIA-NN (v1.8.1) based on the respective FASTA files, whereas MaxDIA employed its native library prediction engine. Crucially, the match between runs (MBR) functionality remained disabled to ensure a standardized assessment of identification performance across all platforms. Analyses using MSFragger-DIA were conducted via the DIA-SpecLib-Quant workflow, with mass tolerances set to 10 ppm for precursors and 20 ppm for fragments.

Quantification assessments utilized precursor-level data across all evaluated platforms. Precursor quantities were extracted from the 'Precursor.Quantity' column in DIA-NN results and the 'Intensity' column within evidence.txt files of MaxDIA. For protein-level quantification, a uniform approach involved calculating the arithmetic mean of constituent peptide intensities following the removal of outliers.

All other parameters for benchmarking software tools were maintained at their default settings, and comprehensive configuration files are provided in the Supporting Information.

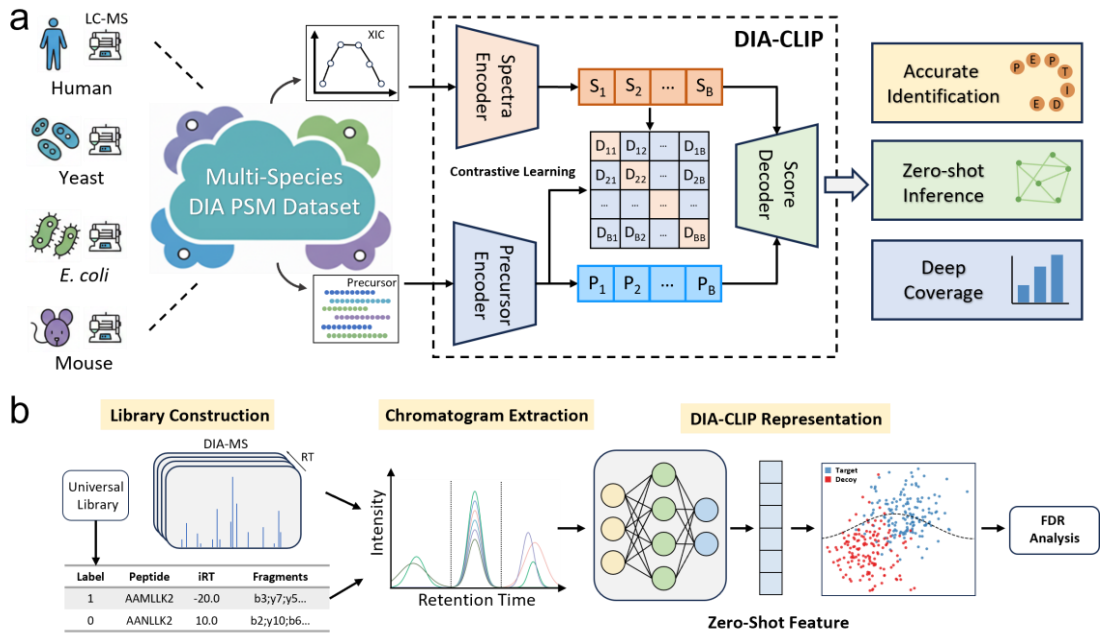

**Figure1. Overview of DIA-CLIP.** a. Illustration of the supervised contrastive pre-training strategy. DIA-CLIP leverages large-scale, diverse PSM dataset to learn robust and generalizable feature representations. By aligning precursors and corresponding XIC signals into a shared latent space, DIA-CLIP facilitates deeper and more precise peptide identification with zero-shot inference. b. Schematic representation of the DIA-CLIP-integrated analysis pipeline. DIA-CLIP is seamlessly incorporated into the standard DIA-MS workflow, specifically at the PSM re-scoring stage, to enable high-throughput and scalable proteomic profiling across diverse practical scenarios.

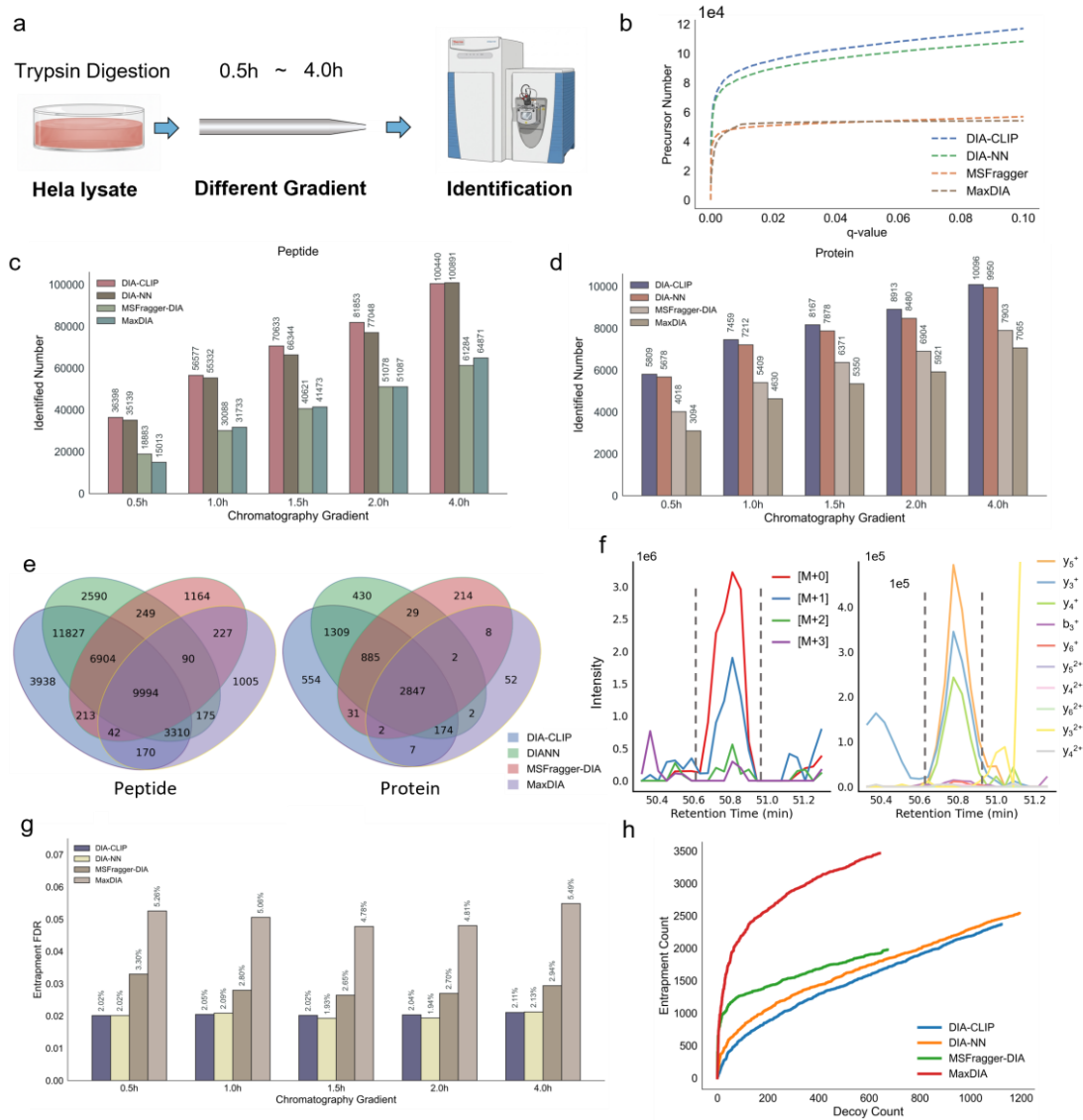

**Figure 2. Benchmark evaluation of DIA-CLIP for Hela cell lysates DIA-MS data.** a. Schematic of the benchmarking experimental design. The dataset utilizes HeLa cell lysates across five distinct chromatographic gradients (30, 60, 90, 120, and 240min). b. FDR curve of precursor identification across various search engines at 90-minute LC gradient. c-d. Benchmarking results for peptide (c) and protein (d) identification across different LC gradient lengths. e. Venn diagram illustrating the overlap and unique counts of peptide and protein between DIA-CLIP and other software tools under 30-minute LC gradient. f. Representative XICs of target precursor that is uniquely identified by DIA-CLIP but missed by DIA-NN. The high peak-shape correlation across precursor and fragments provides evidence for the reliability of identifications. g. Evaluation of entrapment precursor FDR across different LC gradient lengths with decoy FDR < 0.01. h. Correlation between decoy and entrapment precursor counts at 60-minute LC gradient.

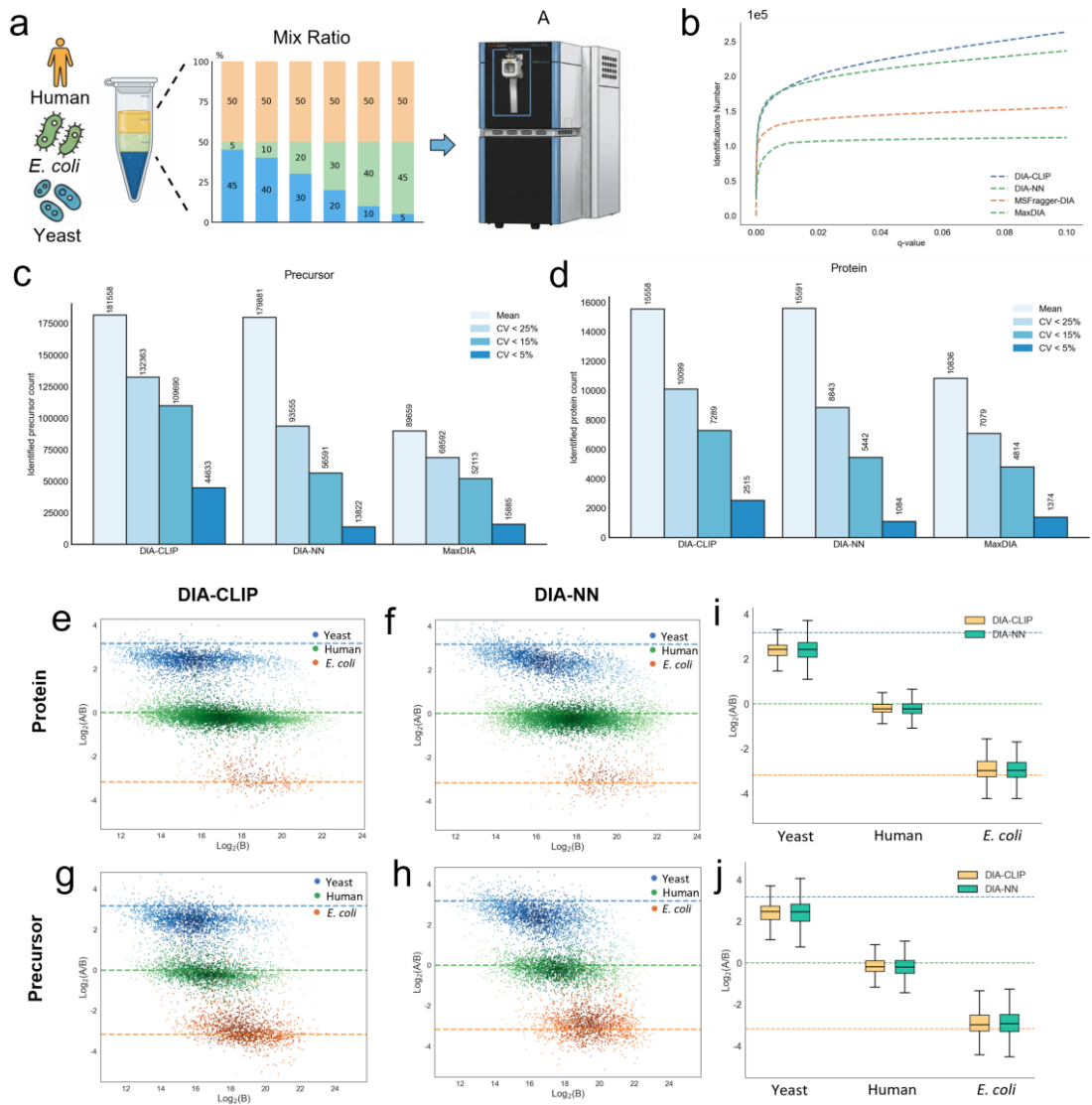

**Figure 3. Benchmarking Results of Multi-species Samples**. a. Experimental design of benchmarking dataset. The samples are generated by mixing known, varying ratios of cell lysates with *Homo sapiens* (human), *Escherichia coli* (*E. coli*), and *Saccharomyces cerevisiae* (yeast). b. FDR curve for precursor identification of various search engines. The analysis is performed on sample with Yeast: Human: *E. coli*=5: 50: 45. c-d. Illustration of identification count of precursors (c) and proteins (d) within various coefficient of variation (CV) on sample with Yeast: Human: *E. coli*=5: 50: 45.. e-h. Scatter plots of relative quantification for DIA-CLIP (e & g) and DIA-NN (f & h) at the precursor level (e-f) and protein level (g-h). The specific ratio change is calculated as Yeast: Human: *E. coli* = 5: 50: 45 relative to Yeast: Human: *E. coli* = 45: 50: 5. The colored dashed lines represent the expected theoretical $\log_2(A/B)$ values of Human (green), Yeast (blue), and *E. coli* (orange). i-j. Box Plots of relative quantification for DIA-CLIP and DIA-NN at precursor level (i) and protein level (j).

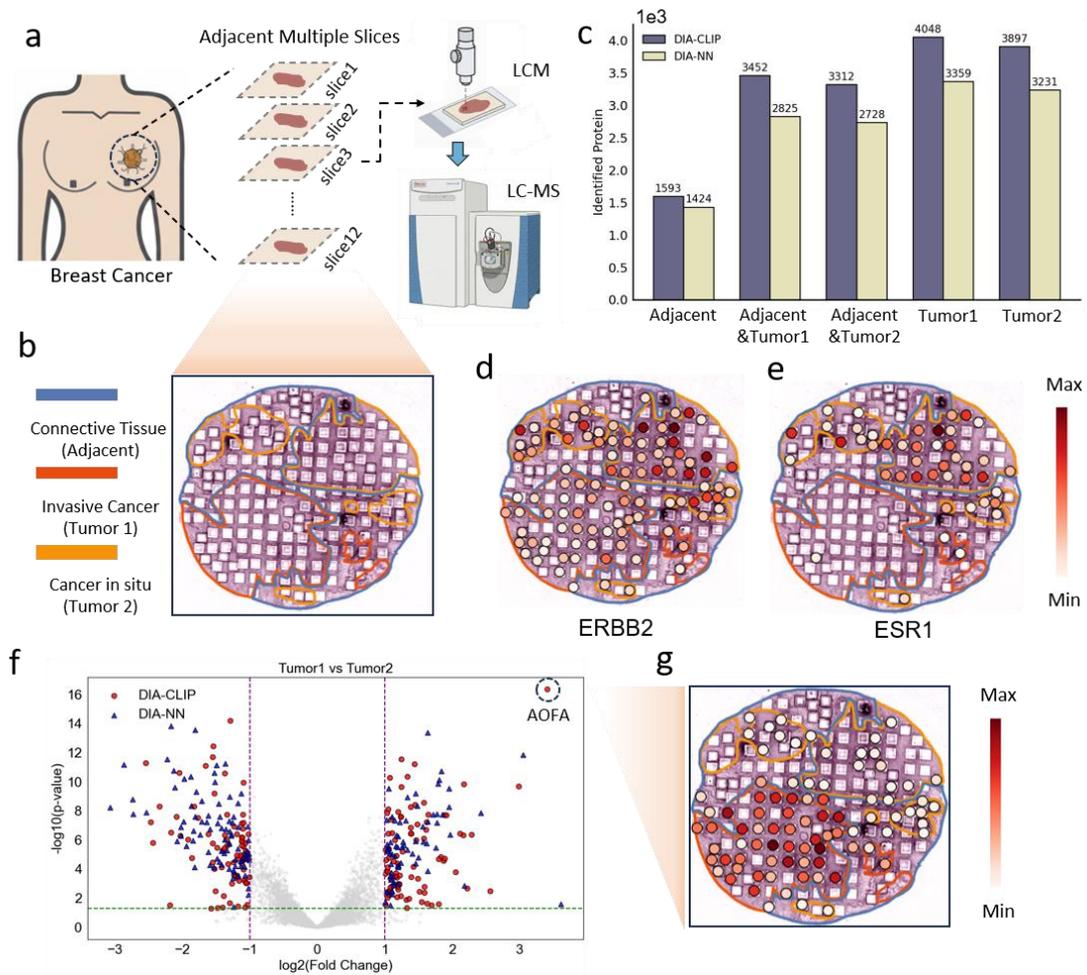

**Figure 4. Application of DIA-CLIP in Spatial Proteomics.** a. Schematic representation illustrating experimental setup of the dataset. One selected breast cancer tissue slice is subjected to laser capture microdissection (LCM), followed by liquid chromatography-mass spectrometry (LC-MS) analysis. b. LCM sampling sites and pathological annotations of breast cancer based on hematoxylin and eosin (H&E) staining images. White squares indicate the LCM sampled areas and colored lines represent the corresponding pathological annotations. c. Comparison of protein identification counts across different pathological regions. Only proteins identified in more than 50% of the tissue sections (LCM samples) are included in the statistics. d-e. Spatial distribution map of epithelial growth factor receptor 2 (ERBB2; d) and estrogen receptor 1 (ESR1; e). Both are identified and quantified by the DIA-CLIP. f. Volcano plots comparing differentially expressed proteins (DEPs) identified by DIA-CLIP and DIA-NN between the two distinct breast cancer pathological regions. g. Spatial distribution map of the representative DEP (AOFA protein) uniquely identified by DIA-CLIP between two distinct breast cancer pathological regions.

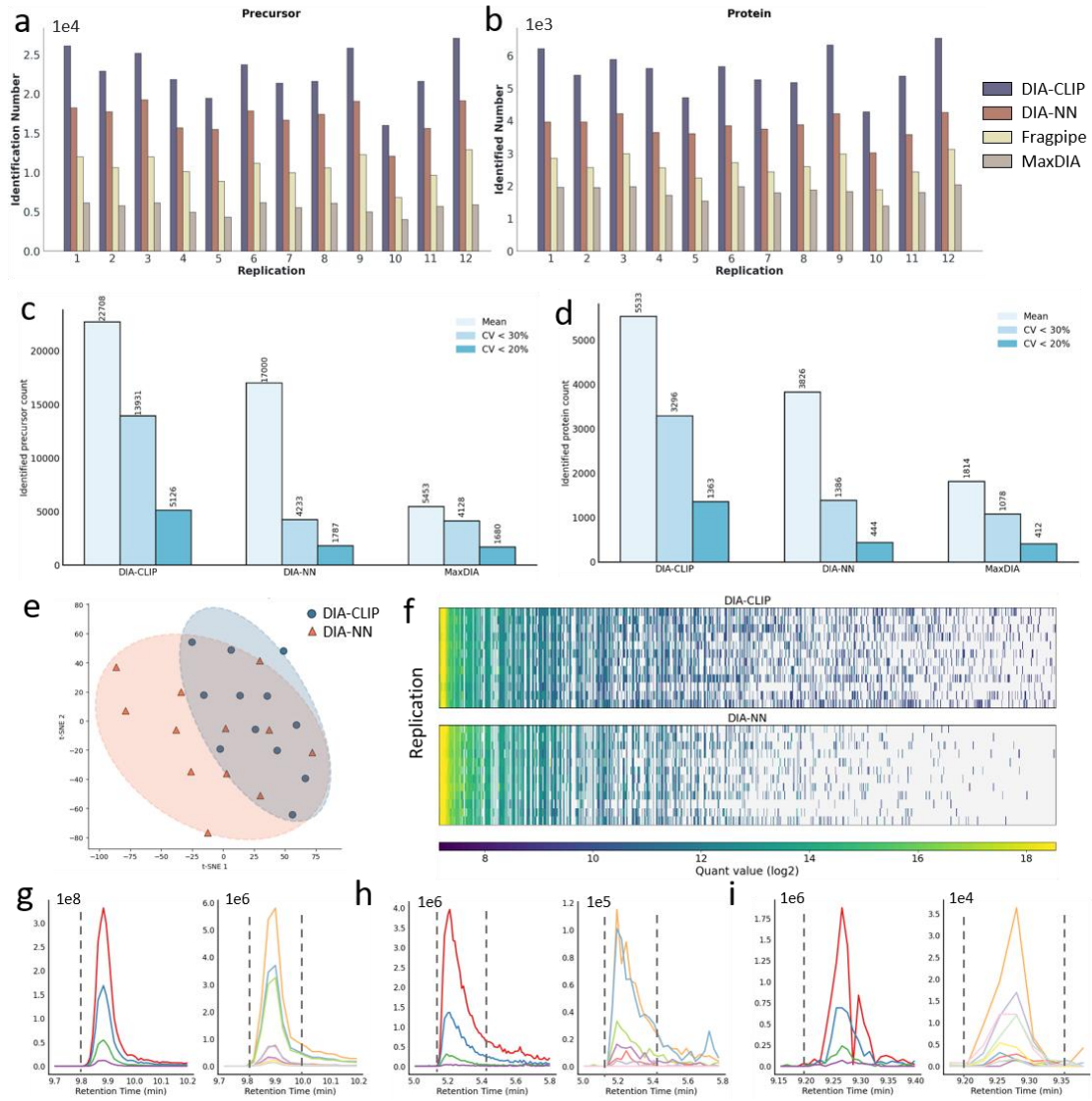

**Figure 5. Application of DIA-CLIP in Single Cell Proteomics.** a-b. Benchmarking of precursors (a) and proteins (b) identification counts across biological replication. c-d. Quantitative analysis of identified precursors (c) and proteins (d) within different CV across replications. e. t-SNE visualization of single-cell proteomes analyzed by DIA-CLIP and DIA-NN. DIA-CLIP reduced mean Euclidean distance from 51.366 (DIA-NN) to 41.015. f. Comparative analysis of data completeness across biological replicates. Protein abundance was ranked from high to low (left to right). g-i. Representative XICs of target precursors that were uniquely identified by DIA-CLIP but missed by DIA-NN.